%% file: PaperForReview.tex
\documentclass[10pt,twocolumn,letterpaper]{article}

\usepackage{cvpr}              

\usepackage{graphicx}
\usepackage{amsmath}
\usepackage{amssymb}
\usepackage{booktabs}

\usepackage{color}
\usepackage{tabularray}
\usepackage{multirow}
\usepackage{colortbl}
\usepackage{pifont}
\usepackage[pagebackref,breaklinks,colorlinks]{hyperref}

\usepackage[capitalize]{cleveref}
\crefname{section}{Sec.}{Secs.}
\Crefname{section}{Section}{Sections}
\Crefname{table}{Table}{Tables}
\crefname{table}{Tab.}{Tabs.}

\def\cvprPaperID{*****} 
\def\confName{CVPR}
\def\confYear{2022}

\begin{document}

\title{Sticker820K: Empowering Interactive Retrieval with Stickers}

\author{Sijie Zhao$^1$\quad Yixiao Ge$^1$\quad Zhongang Qi$^2$\quad Lin Song$^1$\quad Xiaohan Ding$^1$\quad Zehua Xie$^2$\quad Ying Shan$^{1,2}$\\
$^1$Tencent AI Lab\qquad
$^2$IPS Search, Tencent PCG \\
{\tt\small \{sijiezhao, yixiaoge, zhongangqi, ronnysong, xiaohanding, zehuaxie, yingsshan\}@tencent.com}}
\maketitle

\input{0-abstrct}

\input{1-introduction}

\input{2-related_work}

\input{3-dataset}

\input{4-sticker_clip}

\input{5-sticker_llm}

\input{6-conclusion}

\clearpage
{\small
\bibliographystyle{ieee_fullname}
\bibliography{egbib}
}

\end{document}

%% file: 0-abstrct.tex
\begin{abstract}

Stickers have become a ubiquitous part of modern-day communication, conveying complex emotions through visual imagery. To facilitate the development of more powerful algorithms for analyzing stickers, we propose a large-scale Chinese sticker dataset, namely Sticker820K, which consists of 820k image-text pairs. Each sticker has rich and high-quality textual annotations, including descriptions, optical characters, emotional labels, and style classifications. Although vision-language tasks in the domain of natural images have been well studied, directly applying the those models, such as CLIP, to sticker data is not an optimal solution due to the discrepant nature between natural and emotive image data. Therefore, we propose StickerCLIP as a benchmark model on the Sticker820K dataset. For the text-to-image retrieval task, our StickerCLIP demonstrates strong superiority over the CLIP, which achieves an absolute gain of 66.0\% in mean recall on the Sticker820K test set. Additionally, we endeavor to extend the recently popularized LLM by means of prompt tuning, integrating its ability for sticker retrieval and allowing users to retrieve stickers through instructions. We validate the feasibility of this method, demonstrating the immense potential of prompt tuning in expanding LLM abilities while not affecting the quality of upstream tasks. Project page: \href{https://github.com/sijeh/Sticker820K}{https://github.com/sijeh/Sticker820K}.
\end{abstract}

%% file: 1-introduction.tex
\section{Introduction}

Stickers have become increasingly popular in today's digital world. They are an efficient and fun way of conveying emotions and ideas in one's messages or social media posts. The increasing usage of stickers has led to a growing need for better understanding and retrieval of these visual elements. 

A significant difference between stickers and natural images is that stickers often have abstract shapes and icons, rich emotional expressions, and may also include optical characters on them. This makes it challenging to directly apply existing algorithms like CLIP \cite{radford2021learning,yang2022chinese} to sticker understanding. Although the understanding of natural images has been extensively studied, there has not been a deep exploration in the field of stickers. One reason for this is the lack of relevant datasets and benchmarks. Therefore, in this paper, we propose a high-quality Chinese multimodal emoticon dataset, namely Sticker820K, which consists of 820k image-text pairs. Each sticker has rich and high-quality textual annotations, including descriptions, optical characters, emotional labels, and style classifications. This study aims to promote not only the understanding of objective content in images, but also a better understanding of emotions conveyed by images.

To comprehend the semantic and emotional cues of stickers, we have introduced two algorithms based on this dataset: 1) StickerCLIP. Due to its success in multimodal feature alignment, CLIP has become a foundational model for numerous vision language tasks. Building upon this model, our StickerCLIP algorithm further explores the alignment of visual and textual features with emotional cues and artificial painting, enabling it to better accomplish tasks such as sticker retrieval. 2) StickerLLM. Recently, large language models (LLM), such as ChatGPT \cite{ouyang2022training} and LLaMA \cite{touvron2023llama}, have achieved enormous success in the field of natural language processing, garnering widespread attention and research. Additionally, VisualGPT \cite{wu2023visual} and HuggingGPT \cite{shen2023hugginggpt} have further expanded the capabilities of LLM in using other tools. Works such as BLIP2 \cite{li2023blip} and FROMAGe \cite{koh2023grounding} have tried to extend frozen LLM to multi-modal tasks.  In this paper, we explored the possibility of integrating other tools into LLM without compromising its language processing abilities, taking sticker retrieval as our starting point. Previous methods implemented tool integration through prompt engineering, where text context would increase with the number of tools, resulting in additional computational costs during inference. StickerLLM aims to integrate tools through prompt tuning in the form of additional tokens without increasing text context length.

Our main contributions are as follows:
\begin{enumerate}
    \item We collect Sticker820K, a large scale Chinese sticker dataset which consists of 820k image-text pairs. Each sticker has rich and high-quality textual annotations, including descriptions, optical characters, emotional labels, and style classifications. To our knowledge, this is the largest Chinese sticker dataset as so far.
    \item We introduce StickerCLIP, a benchmark model trained on Sticker820K dataset, which enables more effective alignment between stickers and texts.
    \item We propose StickerLLM, a model that enhances the ability of LLM to perform sticker retrieval without affecting upstream tasks. This is achieved by introducing special tokens into the frozen LLM and updating only the embeddings of these tokens.
\end{enumerate}

%% file: 2-related_work.tex
\section{Related Works}

\subsection{Vision-Language Datasets}
\textbf{Natural image-text pair datasets.} The progress in the Vision-Language (VL) tasks owes much to the emergence of various multimodal datasets in recent years. At present, the available vision-language datasets mainly consist of image-text pairs that are related to natural images. Some of the smaller datasets in this category are Flickr30k \cite{plummer2015flickr30k}, COCO-Captions \cite{lin2014microsoft}, Visual Genome \cite{krishna2017visual}, and VQAv2 \cite{goyal2017making}. On the other hand, larger datasets, such as CC3M \cite{sharma2018conceptual}, CC12M \cite{changpinyo2021conceptual}, YFCC100M \cite{thomee2016yfcc100m}, Wukong \cite{gu2022wukong}, LAION-400M \cite{schuhmann2021laion}, and LAION-5B \cite{schuhmann2022laion}, have millions of samples. These datasets have significantly facilitated the pre-training of VL models, making them highly effective in downstream tasks through zero-shot learning. However, stickers, as a kind of visual image, exhibit features vastly different than those of natural photos. For instance, even if they portray similar visual characteristics, stickers may express very contrasting emotions, making them unfit for use in the same context. Additionally, stickers contain hand-drawn abstract patterns and non-negligible optical characters.

\textbf{Emotional vision-language datasets.} Stickers find widespread usage in social media applications, as they can significantly improve the communication experience. Datasets relevant to stickers can be classified into two domains: dialogue and sentiment analysis. Among these, the first-class datasets such as DSTC10-MOD \cite{fei2021towards} and SRS \cite{gao2020learning} are mainly employed in recommending emoticons based on the context of the conversation. On the other hand, the second-class datasets, such as SER30K \cite{liu2022ser30k}, MELD \cite{poria2018meld}, and HatefulMemes \cite{kiela2020hateful}, are used to analyze and classify the emotions conveyed by emoticons. It is worth noting that these datasets lack detailed descriptions of stickers. For example, the emoticon dialogue dataset only includes context related to emoticons, while the sentiment analysis dataset contains only annotations of emotion categories. In contrast, our proposed dataset encompasses semantic descriptions of emoticons, their emotional categories, style categories, and optical character information. The primary objective of this dataset is to facilitate further research in the field of emoticon understanding.

\subsection{Vision-Language Methods}
In recent years, the realm of universal methodologies has witnessed extensive research on vision-language tasks, which can be broadly categorized into cross-modal comprehension and cross-modal generation. Within the domain of cross-modal understanding, a critical point lies in aligning visual and textual features. Techniques such as CLIP \cite{radford2021learning} and ALIGN \cite{jia2021scaling} employ a dual-tower model to separately extract these features, aligning them through a global contrastive learning approach. UNITER \cite{chen2020uniter}, on the other hand, utilizes a multimodal encoder to simultaneously extract visual and textual characteristics. Building upon the foundation of contrastive learning, ALBEF \cite{li2021align} introduces image-text matching and the mask language model. FILIP \cite{yao2021filip} harnesses the granular representational capability of image patches and textual words to further refine feature alignment. We observed that CLIP exhibit suboptimal performance within the Sticker820K dataset, primarily due to a substantial gap between pretraining data and sticker data. Thus we propose StickerCLIP as a benchmark model to deal with this problem. As for cross-modal generation tasks, BLIP2 \cite{li2023blip}, FROMAGe \cite{koh2023grounding}, Mini-GPT4 \cite{zhu2023minigpt}, LLaVA \cite{liu2023visual} extend pretrained LLM to multi-modal tasks, such as generating text based on visual information. VisualGPT \cite{wu2023visual}, HuggingGPT \cite{shen2023hugginggpt}, GPT4Tools \cite{yang2023gpt4tools} utilizes prompt engineering to equip LLM with the ability to use external tools, which greatly increases the context length, leading to higher computational requirements and longer response time for users. Our StickerLLM add special tokens to LLM and optimize their embeddings through prompt tuning, which expands the capability of tool without the need for additional context.


%% file: 3-dataset.tex
\section{Sticker820K Dataset}
\subsection{Data Construction}

\begin{figure*}[htbp]
  \centering
  \includegraphics[width=\textwidth]{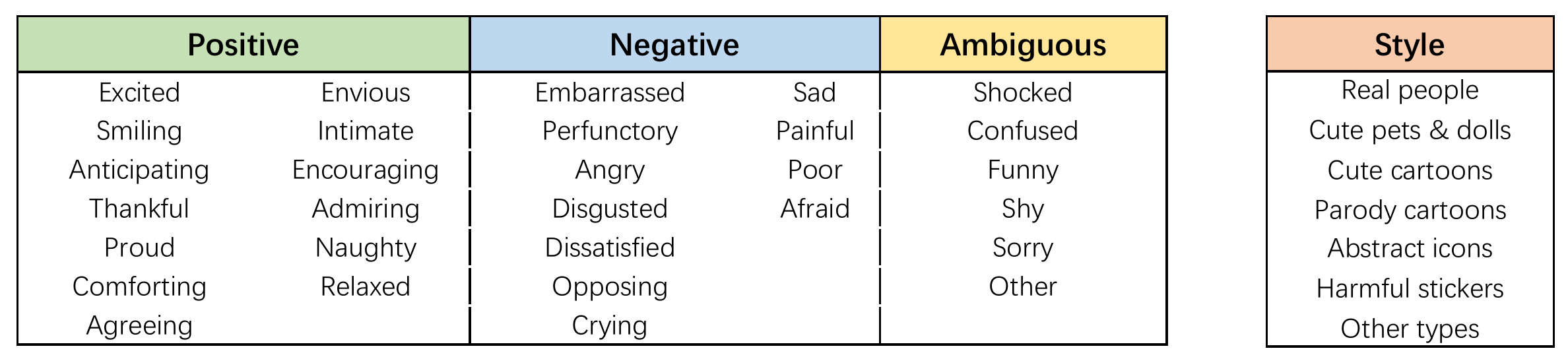}
  \caption{\textbf{Left}: Predefined emotional categories for the stickers, which can be roughly divided into positive, negative, and ambiguous. \textbf{Right}: Predefined style categories for the stickers. }
  \label{fig:ann_emo_style}
\end{figure*}

In this paper, we present a dataset composed of 820k image-text pairs, initially collected and cleansed from search engines such as Sogou and Baidu. This dataset encompasses static image formats like PNG and JPG, as well as dynamic GIF images. We established annotation guidelines and released annotation tasks on a crowdsourcing platform.

For each image, Our Sticker820K provide the following information: 1) A detailed content description, capturing the essence of the stickers. 2) An emotional category, based on studies like GoEmotions \cite{demszky2020goemotions}, combined with unique emotional expressions in Chinese, resulting in a rich selection of 30 categories - 15 positive, 11 negative, and 4 ambiguous emotions. The detailed emotional categories is shown in Fig. \ref{fig:ann_emo_style} (left). Since a sticker can sometimes convey more than one emotion, annotators were asked to consider potential social conversations when labeling emotions. 3) Image style, classified into five categories based on observations, as shown in Fig. \ref{fig:ann_emo_style} (right). We categorized cute pets and plush toys into same class, because of their similar visual feature, and we separately distinguished harmful expressions to prevent toxic results in potential applications. 5) Optical characters within the image, as the text found in memes carries significant meaning, with different texts on the same image potentially conveying distinct implications. Utilizing Optical Character Recognition (OCR) tools, we extracted text from these images. 

We partition Sticker820K into a training set and test set, with the training set comprising 740,017 image-text pairs and the test set containing 82,225 image-text pairs.

\subsection{Dataset Characteristics}

\begin{figure*}[htbp]
  \centering
  \includegraphics[width=\textwidth]{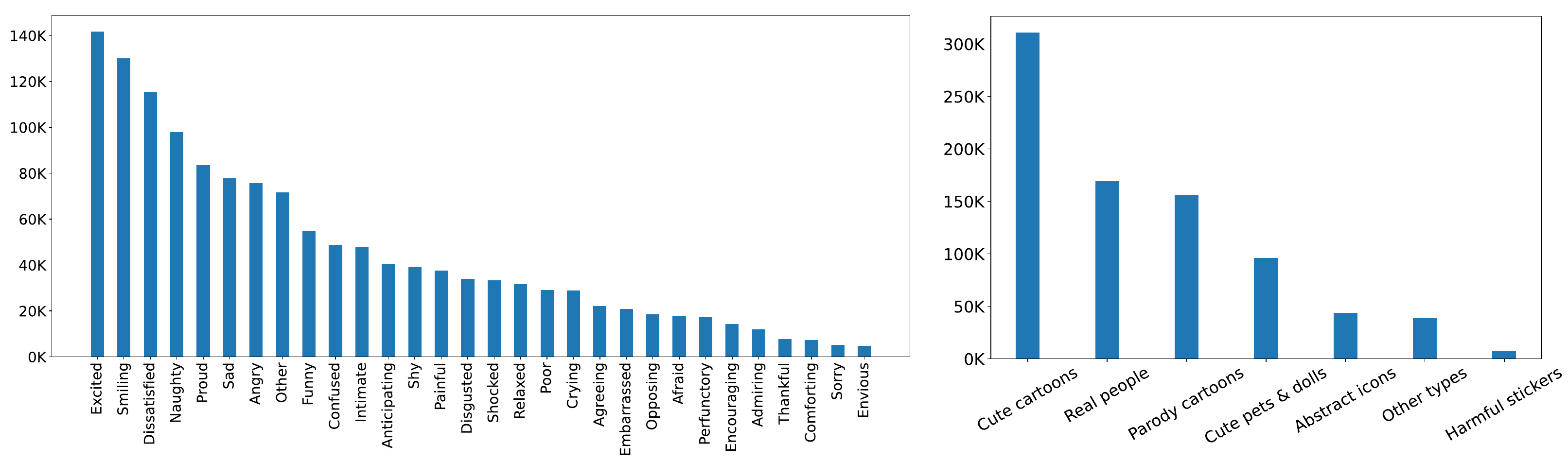}
  \caption{Left: frequency of emotion labels in the proposed datasets. Right:  frequency of style labels in the proposed datasets}
  \label{fig:stat_emo_style}
\end{figure*}

We have presented in the Fig. \ref{fig:stat_emo_style} (left) the frequency of emotion labels for the sticker data, wherein it can be observed that the positive category encompasses a majority of the samples, whilst the negative category is comparatively sparse. In the Fig. \ref{fig:stat_emo_style} (right), we have shown the style labels for the sticker data, and it can be noted that Cute Cartoons stands as the most predominant component, which also aligns with our actual usage patterns during social communication.

\begin{figure*}[htbp]
  \centering
  \includegraphics[width=\linewidth]{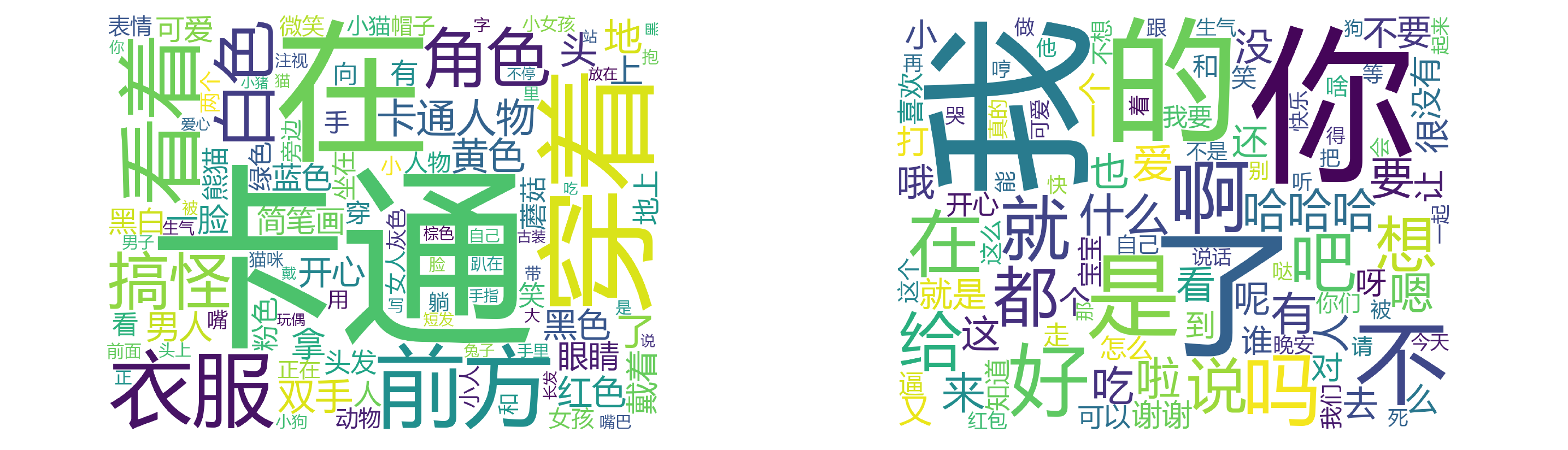}
  \caption{Left: Word clound distribution of the content description in the Sticker820K. Right: Word cloud distribution of optical characters in the Sticker820K.}
  \label{fig:word_clound}
\end{figure*}

\begin{figure}
\centering
\includegraphics[width=0.47\textwidth]{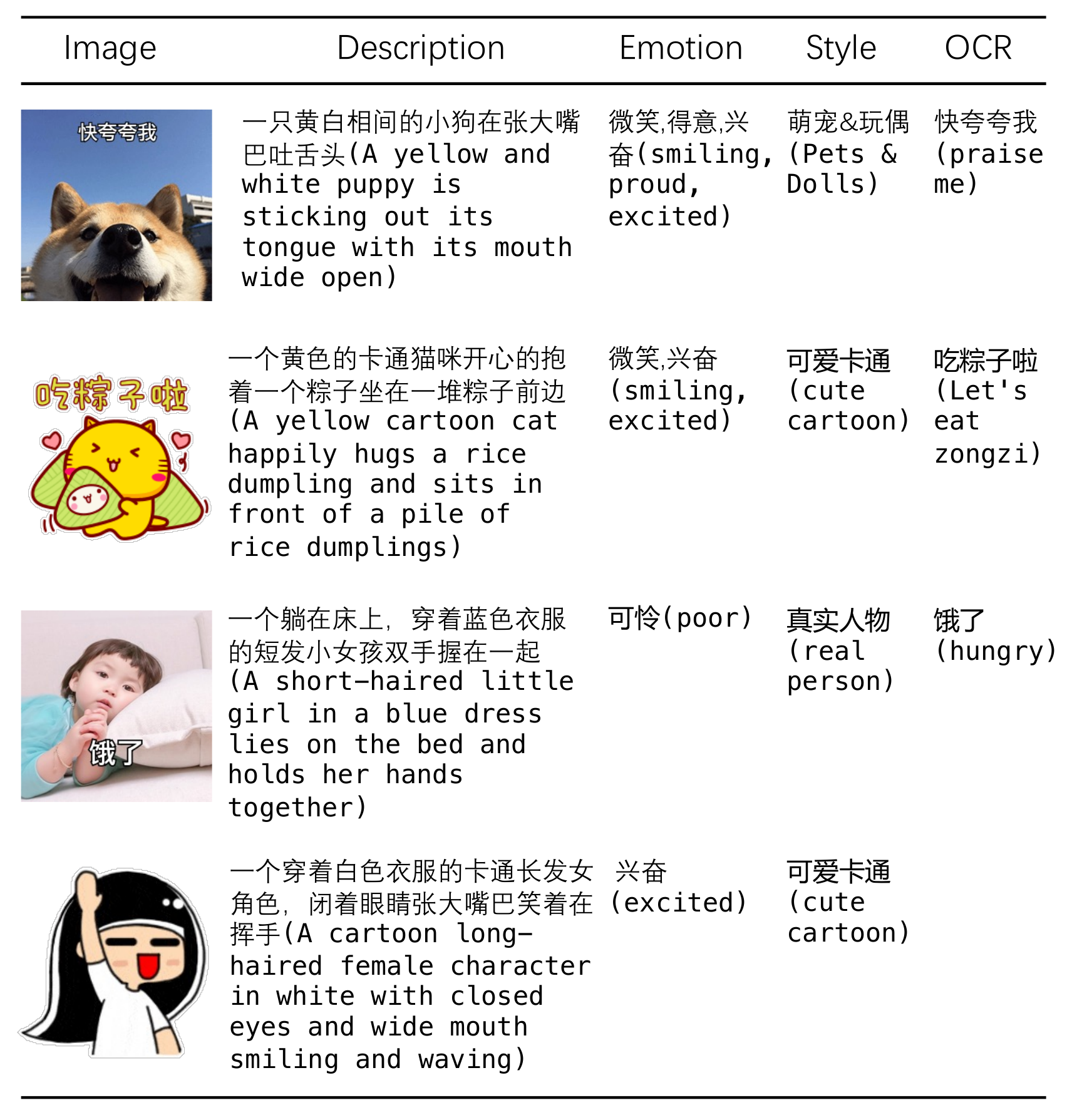}
\caption{Data examples of Sticker820K.}
\label{fig:data_example}
\end{figure}

As shown in Fig. \ref{fig:data_example}, The dataset we propose contains rich textual information. It comprises not only manually annotated escription, but also text extracted from images via optical character recognition. In Fig. \ref{fig:word_clound}, we demonstrate the word cloud distributions of both types of textual information in this dataset. The left side displays the word cloud distribution of content description, which shows that the most common words include ``cartoon",``wearing", ``looking", ``character", and ``front". On the right side, the word cloud distribution of optical characters detected is shown. It differs a lot from the caption word cloud, with the most common words being ``me", ``you", and so on.

%% file: 4-sticker_clip.tex
\section{StickerCLIP}

\begin{figure*}[t]
  \centering
  \includegraphics[width=\textwidth]{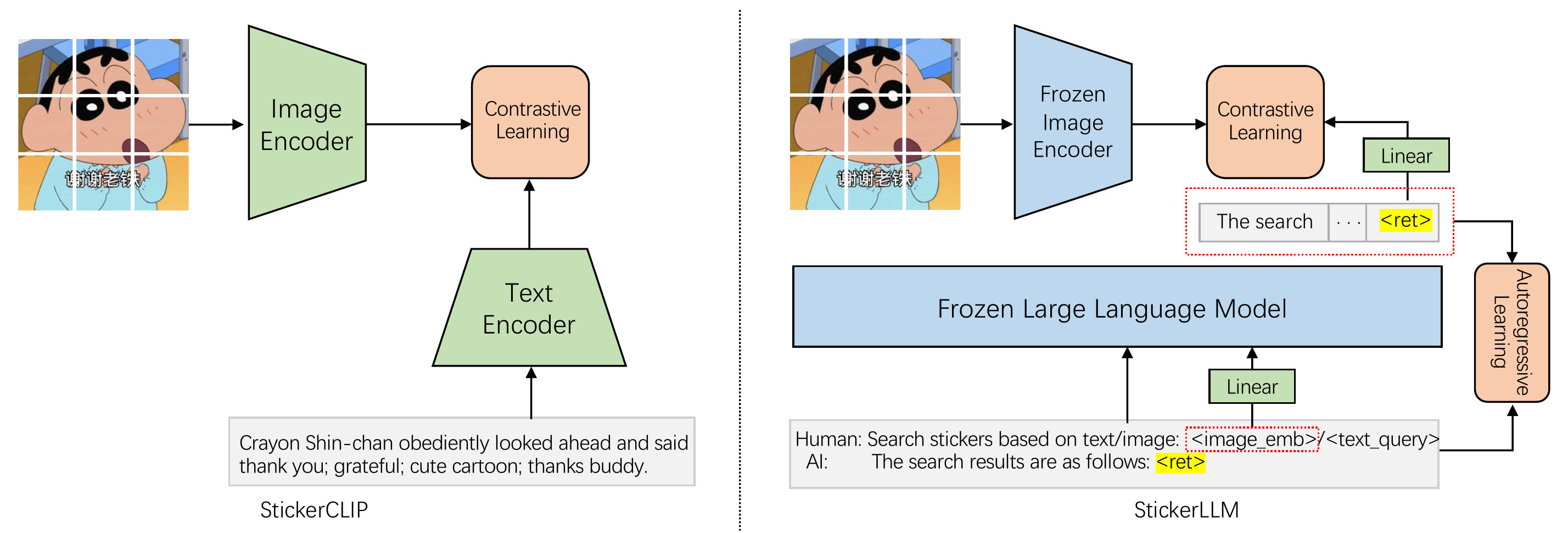}
  \caption{Overview of StickerCLIP (left) and StickerLLM (right). Left: we finetune CLIP on Sticker820K to align features between stickers and texts. Right: we use Frozen LLM and image encoder to construct StickerLLM which extends image as input and retrieves stickers by following user's instruction. }
  \label{fig:framework}
\end{figure*}

\subsection{Methodology}
Based on the Sticker820K, we propose StickerCLIP to align the stickers and text. Our StickerCLIP is constructed on ChineseCLIP \cite{yang2022chinese} which comprises a visual encoder $E_{\theta}$ and a text encoder $E_{\phi}$. Given a caption $T$ and its paired image $I$,  $E_{\theta}(I) \in \mathbb{R}^{d}$ and $E_{\phi}(T) \in \mathbb{R}^{d}$ represent the outputs of the visual encoder and text encoder, respectively. 
In order to align the text and visual features, CLIP employs InfoNCE loss \cite{oord2018representation} for text-to-image(t2i) and image-to-text(i2t) retrieval over a batch of N image-text pairs $(T_i,I_i)$, which can be formulated as follows:
\begin{align}
\mathcal{L}_{\mathrm{t} 2 \mathrm{i}} & =-\frac{1}{N} \sum_{i=1}^N\left(\log \frac{\exp \left(\operatorname{sim}\left(T_i, I_i\right) / \tau\right)}{\sum_{j=1}^N \exp \left(\operatorname{sim}\left(T_i, I_j\right) / \tau\right)}\right) \label{func:info_nce_t2i}\\
\mathcal{L}_{\mathrm{i} 2 \mathrm{t}} & =-\frac{1}{N} \sum_{i=1}^N\left(\log \frac{\exp \left(\operatorname{sim}\left(I_i, T_i\right) / \tau\right)}{\sum_{j=1}^N \exp \left(\operatorname{sim}\left(I_i, T_j\right) / \tau\right)}\right) \label{func:info_nce_i2t}
\end{align}
In the above formula, $\operatorname{sim}(I,T)$ represents the cosine similarity between the feature vectors of the image and the caption. $\tau$ is a learnable temperature parameter. The total loss is:
\begin{equation}
    \mathcal{L}_{total} = \mathcal{L}_{\mathrm{t} 2 \mathrm{i}} + \mathcal{L}_{\mathrm{i} 2 \mathrm{t}}
\end{equation}
We concatenate different type of textual information and use it as input for the text encoder. Since the stickers we collected not only include static images, such as JPG and PNG, but also animated images, such as GIF, while the pre-trained CLIP model only takes static images as input. Therefore, we extract multiple frames of an animated image as input for the visual encoder and calculate the average vector along different frames. 

\subsection{Implementation Details}
In this experiment, We train StickerCLIP on Sticker820K, the batch size equates to 2,048, the learning rate to 2e-5, and the weight decay to 1e-2. We employ the AdamW \cite{loshchilov2017decoupled} optimizer in tandem with a Cosine learning rate schedule, designate the content length of the text encoder as 64, and due to the inclusion of animated images, we extract features from the first, middle, and last frames of a sticker as input for the visual encoder, subsequently averaging these features to obtain the final aligned embedding. 

\subsection{Performance}

\begin{table*}
\centering
\setlength\tabcolsep{9pt}
\begin{tabular}{llccccccccc} 
\toprule
\multicolumn{1}{c}{\multirow{2}{*}{Model}} & \multicolumn{1}{c}{\multirow{2}{*}{Method}} & \multicolumn{1}{l}{\multirow{2}{*}{Params}} & \multicolumn{4}{c}{Image-to-Text Retrieval} & \multicolumn{4}{c}{Text-to-Image Retrieval}  \\ 
\cline{4-11}
\multicolumn{1}{c}{}                       & \multicolumn{1}{c}{}                        & \multicolumn{1}{l}{}                        & R@1  & R@5  & R@10 & MR                     & R@1  & R@5  & R@10 & MR                      \\ 
\hline
\multirow{3}{*}{RN50}                      & Zero-shot                                   & 0M                                          & 5.8  & 12.8 & 17.0 & 11.9                   & 3.4  & 7.6  & 10.3 & 7.1                     \\
                                           & Freeze-vision                               & 39M                                         & 43.3 & 61   & 66.5 & 56.9                   & 33.0 & 52.3 & 59.3 & 48.2                    \\
                                           & StickerCLIP                                 & 77M                                         & 59.0 & 76.0 & 80.3 & 71.8                   & 57.9 & 75.4 & 79.8 & 71.0                    \\
\multirow{3}{*}{ViT-B}                     & Zero-shot                                   & 0M                                          & 13.3 & 25.8 & 32.1 & 23.8                   & 8.8  & 18.1 & 23.0 & 16.7                    \\
                                           & Freeze-vision                               & 102M                                        & 58.8 & 77.1 & 81.9 & 72.6                   & 52.4 & 72.9 & 78.5 & 67.9                    \\
                                           & StickerCLIP                                 & 188M                                        & 71.4 & 87.7 & 91.2 & 83.4                   & 71.3 & 86.9 & 90   & 82.7                    \\
\bottomrule
\end{tabular}
\caption{Experimental results of various methods in text-to-image retrieval and image-to-text retrieval.}
\label{tab:sticker_clip_eval}
\end{table*}

We conducted experiments on two distinct model sizes, RN50 and ViT-B, comparing the following three methods: 1) Zero-shot, where we evaluated the pre-trained ChineseCLIP; 2) Frozen-vision, in which we froze the visual encoder within the CLIP and fine-tuned it on the Sticker820K training set; and 3) StickerCLIP, which involved fine-tuning all parameters on the Sticker820K training set. We assessed the performance on the Sticker820K test set and reported Recall@K (recall of the top K candidates) with K values of 1, 5, 10, and the Mean Recall (MR) for both image-to-text and text-to-image retrieval.

Tab. \ref{tab:sticker_clip_eval} displays the performance of various models and methods on the test set. Generally speaking, StickerCLIP outperforms the other two methods, achieving the best overall performance. For instance, under the same model of ViT-B, StickerCLIP exhibits a 59.6\% and 66.0\% improvement respectively over Zero-shot in the Mean Recall (MR) metric for Image-to-Text Retrieval and Text-to-Image Retrieval, and a 10.8\% and 14.8\% increase respectively in comparison to Frozen-vision. The inferior zero-shot performance of CLIP on StickerCLIP suggests that the sticker dataset exhibits significant differences in data features compared to the natural image-text pairs used for pre-training. Specifically, on the visual side, the Sticker820K dataset contains numerous hand-drawn images, such as cartoons, while on the textual side, Sticker820K encompasses a rich assortment of emotive vocabulary. Consequently, a considerable performance gap persists between StickerCLIP and models that fine-tune the text encoder only.

\subsection{Qualitative Results}

\begin{figure}
\centering
\includegraphics[width=0.47\textwidth]{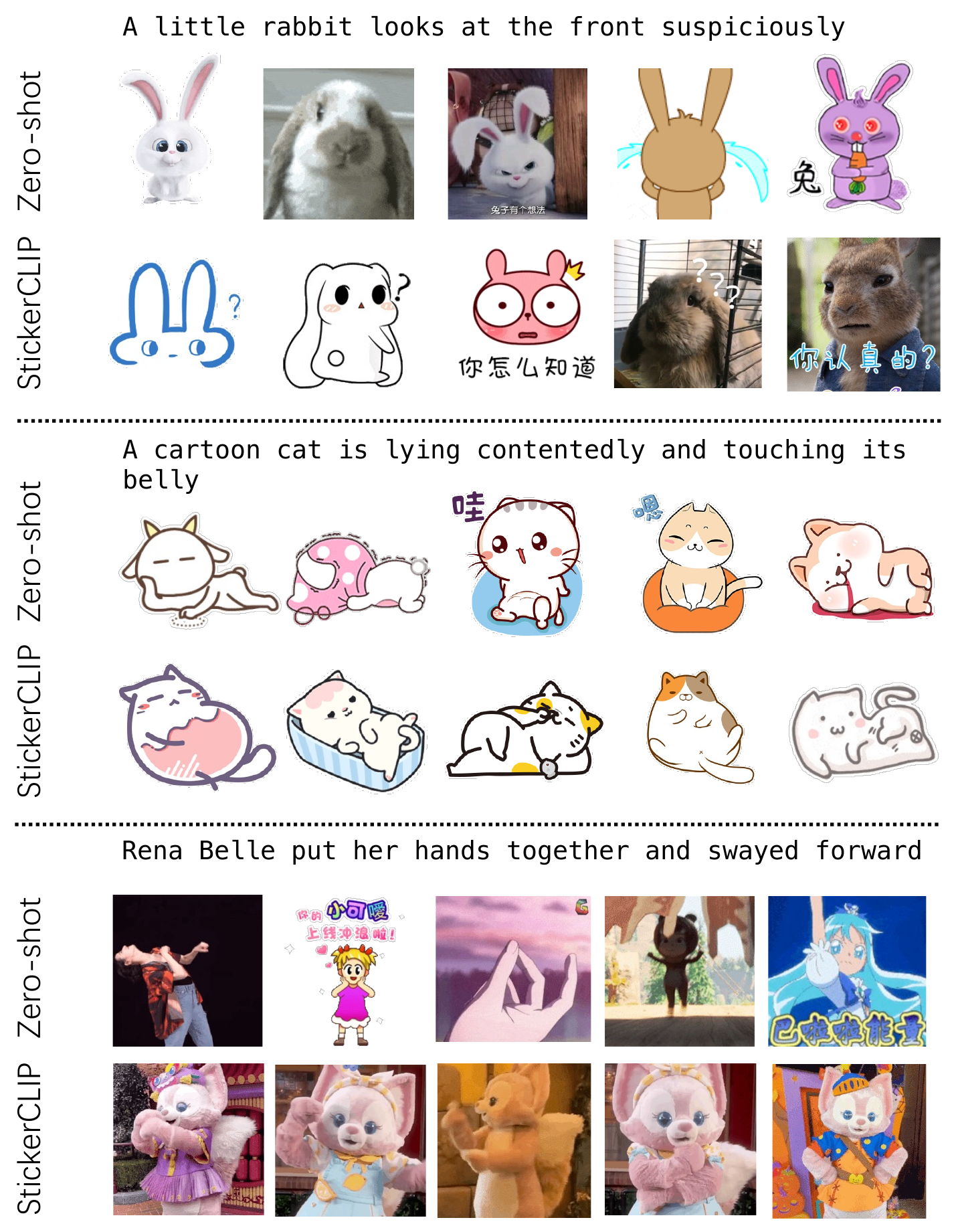}
\caption{Visualization results of StickerCLIP on Sticker820K.}
\label{fig:stickerclip_example}
\end{figure}

\textbf{Visualization of StickerCLIP.} We visualized the retrieval results of StickerCLIP and compared them with zero-shot retrieval results on the Sticker820K dataset, as shown in Fig. \ref{fig:stickerclip_example}. In this comparison, both our StickerCLIP and the Zero-shot model use the ViT-B structure, and we have shown the top 5 results of the same query retrieval. It can be seen that StickerCLIP performs better. For example, in the first example, the results of StickerCLIP can better express the state of confusion, in the second example, most of the results returned by the zero-shot model do not match the theme, in the third example, the Zero-shot model does not understand the meaning of ``Rena Belle".

%% file: 5-sticker_llm.tex
\section{StickerLLM}
Recently, LLM has garnered widespread attention and undergone intensive research due to its powerful text processing capabilities. Researchers have aligned LLM with human intent via instruction tuning and equipped it with the ability to utilize tools through prompt engineering. Furthermore, through expanding frozen LLM, researchers have enabled it to process multimodal information. 

\subsection{Methodology}
In this paper, we further explore the potential to expand the capabilities of LLM with the goal of enabling it to retrieve stickers without disrupting upstream tasks. To achieve this objective, we add special tokens to LLM as representations of the tools. During the training process, we freeze the majority of the parameters in LLM and only finetune parital weights of embedding layer and logistic head corresponding to the position of the new tokens.

As illustrated in Fig. \ref{fig:framework} (right), StickerLLM comprises an LLM and a visual encoder. To effectively extract visual embedding from stickers, we obtain the visual encoder from finetuned StickerCLIP and freeze all its parameters. The LLM not only generates textual responses through auto-regressive modeling, but is also utilized for extracting retrieval embedding. We introduce the $\langle ret \rangle$ token for this purpose.

To enable StickerLLM to retrieve stickers by following human instructions, the framework need to learn two objectives: 1) generating the text responese which incoporates $\langle ret \rangle$ token at an appropriate position based on contextual clues, and 2) empowering the $\langle ret \rangle$  token with the ability to retrieve stickers.

The first objective can be achieved by maximizing the log likelihood of the token sequence, the loss can be factored as a sum of conditional log probabilities:
\begin{equation}
\mathcal{L}_{c}=\sum_{t=1}^t \log p_\theta\left(s_t \mid s_1, \ldots, s_{t-1}\right)
\end{equation}
where $p_\theta$ is the autoregressive LLM and $(s_1, \dots, s_T)$ is the sequence of input tokens. 

To achieve the second objective, we align the LLM's retrieval embedding with visual embedding by contrastive learning. We use a learnable linear layer $\mathbf{W}_t \in \mathbb{R}^{D \times d}$ to map the last hidden state of $\langle ret \rangle$ to $h_{\theta}(s_i)^{T} \mathbf{W}_t \in \mathbb{R}^{d}$, where $h_{\theta}$ is the LLM without logistic head. The contrastive loss is the same with Eq. \ref{func:info_nce_t2i} and Eq. \ref{func:info_nce_i2t}. The total loss can be written as:
\begin{equation}
    \mathcal{L} = \mathcal{L}_c + \lambda (\mathcal{L}_{\mathrm{t2i}} + \mathcal{L}_{\mathrm{i2t}})
    \label{func:loss_stickerllm}
\end{equation}
where $\lambda$ is the scale of contrative loss. 

To achieve the collaborative search using image and text, we endeavored to extract unified retrieval embeddings of visual and textual data through $\langle ret \rangle$. If image is inputted, we initially extract the visual features through the frozen visual encoder of StickerCLIP. Subsequently, we map the aforementioned visual features onto the input of LLM by utilizing a learnable linear layer $\mathcal{W}_c \in \mathbb{R}^{d\times D}$.

Since the visual encoder and LLM is frozen, only the linear layer $\mathbf{W}_t$, $\mathbf{W}_c$, and the embeddings corresponding to the position of the new special tokens receive gradient updates.

\subsection{Implementation Details}
In StickerLLM, the batch size is setting to 288 due to the limited memory, the learning rate and weight decay are setting to 2e-5 and 0. $\lambda$ in Eq. \ref{func:loss_stickerllm} is setting to 1.0. We employ the AdamW \cite{loshchilov2017decoupled} optimizer in tandem with a cosine learning rate schedule. we construct a single-round retrieval dialogue comprising instruction and answer templates, with queries entailing user retrieval instructions and corresponding responses incorporating special tokens for retrieval. The examples of data template is shown in Tab. \ref{tab:template_example}.  We implement ChatGLM-6B \cite{zeng2023glm-130b} as the language model, introducing new special tokens such as $\langle ret \rangle$, $\langle /ret \rangle$, $\langle img \rangle$, $\langle /img \rangle$, and $\langle pret \rangle$, the meaning of which are outlined in the Tab. \ref{tab:special_tokens}. With a 50\% probability, we add $\langle pret \rangle$ to the beginning of the training data as prefix tuning to modulate generated probability distributions. Our StickerLLM is configured to accept both text and image inputs, facilitating text-to-image retrieval, image-to-image retrieval, and unified image-text-to-image retrieval. During data construction, we generate text-to-image retrieval data with a 50\% likelihood, image-to-image retrieval data with a 25\% probability, and since precise joint retrieval data pairing cannot be directly generated, we randomly extract images from the dataset with a 25\% probability to create image-text-to-image retrieval data, ensuring that LLM pays attention to user input text when images are provided. 

\begin{table}
\centering
\begin{tabular}{cc} 
\toprule
Token & Meaning                        \\ 
\hline
$\langle ret \rangle$   & Used for sticker retrieval.     \\
$\langle /ret \rangle$  & End of retrieval token.         \\
$\langle img \rangle$   & Start of image embedding.       \\
$\langle /img \rangle$  & End of image embedding.         \\
$\langle pret \rangle$  & Prefix to modulate generation.  \\
\bottomrule
\end{tabular}
\caption{The meaning of added special tokens in StickerLLM.}
\label{tab:special_tokens}
\end{table}



\begin{table}
\centering
\small
\setlength\tabcolsep{1pt}
\begin{tabular}{c} 
\toprule
Examples
of Instuction Template~ ~ ~~                                   \\ 
\hline
Retrieve
emoticons based on the following text: \{text\}.~ ~ ~~       \\
Find
stickers:
\{text\}.~ ~ ~~                                        \\
\{image\} retrieve stickers based on the
image.~ ~ ~~                 \\
\{image\}
retrieve similar stickers.~ ~ ~~                            \\
\{image\}
combines image and text to find emoticons: \{text\}.~ ~ ~~  \\
\{image\} Find stickers: \{text\}.~ ~ ~~                               \\ 
\hline
Examples
of Answer Template~ ~ ~~                                     \\ 
\hline
The retrieval results are as follows:
$\langle ret \rangle$ $\langle /ret \rangle$.~ ~ ~~                  \\
$\langle ret \rangle$ $\langle /ret \rangle$.~ ~ ~~                                                        \\
\bottomrule
\end{tabular}
\caption{Examples of instruction and answer template.}
\label{tab:template_example}
\end{table}

\subsection{Performance}
The performance of StickerLLM can be divided into two aspects: (1) the ability to use retrieval tools in the correct context, that is, LLM should only produce responses containing the $\langle ret \rangle$ token when the user presents instructions related to retrieving stickers. (2) The retrieval performance of the $\langle ret \rangle$ token in the retrieval task. We designed the following experiments to evaluate StickerLLM.

\textbf{Accuracy of tool selection.} To assess the accuracy of the retrieval tool, we conducted experiments in four scenarios: (a) instructions not related to retrieving stickers. We randomly selected 1000 instruction data from validation set of BELLE \cite{BELLE}, which can be considered unrelated to retrieving stickers. We calculate the percentage of instructions that do not contain the $\langle ret \rangle$ token as the accuracy rate. (b)-(e) Using retrieval instructions in both in-domain and out-of-domain training sets, with and without prefixes, i.e., the $\langle pret \rangle$ token. We consider the instructions involving in the training set as in-domain data. In (b)-(e), We random sample 1000 instruction in the test set of Sticker820K and calculate the percentage of replies containing the $\langle ret \rangle$ token as the accuracy rate.

\begin{table}[htbp]
\centering
\setlength\tabcolsep{5pt}
\begin{tabular}{lccccc} 
\toprule
          & (a)      & (b)    & (c)    & (d)    & (e)     \\ \hline
Dataset   & BELLE      & Sticker    & Sticker    & Sticker    & Sticker     \\ 
In-domain & \ding{55} & \ding{51} & \ding{55} & \ding{51} & \ding{55}  \\
Prefix    & \ding{55} & \ding{55} & \ding{55} & \ding{51} & \ding{51}  \\ 
Accuracy  & 96.7       & 68.1       & 53.6       & 100.0      & 100.0       \\
\bottomrule
\end{tabular}
\caption{Accuracy of tool selection in different settings. We take $\langle ret \rangle$ token as a retrieval tool and $\langle pret \rangle$ as prefix. }
\label{tab:acc_tool_selection}
\end{table}

As shown in Tab. \ref{tab:sticker_clip_eval}, StickerLLM achieves an accuracy rate of 96.7\% on the BELLE test set. This indicates that for most instructions that are not related to sticker retrieval, StickerLLM doesn't executes the retrieval. Furthermore, since we completely freeze the model parameters (except for the embeddings in the positions of added special tokens), the StickerLLM model will yield similar responses as the original model when the special tokens are not generated. In (b)-(c), it is shown that the absence of prefixes does not alter the model's generation distribution, while the original language model is incapable of generating the $\langle ret \rangle$ token. Therefore, StickerLLM only uses the $\langle ret \rangle$ token for retrieval in sticker search instructions with a frequency of 68.1\% and 53.6\%, respectively. In (d)-(e), since we added the $\langle pret \rangle$ token as a prefix with a 50\% probability in the training data, the $\langle pret \rangle$ token has a stronger modulating ability for the subsequent retrieval tools. Therefore, StickerLLM can correctly invoke the retrieval tool, whether the sticker retrieval instructions are involved in the training data or not.

\textbf{Accuracy of Retrieval.} We also evaluated the accuracy of StickerLLM in retrieving stickers. During training, we constructed templates that unify text and image as input for retrieval instructions. Therefore, we report Recall@K (recall of the top K candidates) with K values of 1, 5, 10, and the Mean Recall (MR) for both image-to-image (I2I) and text-to-image (T2I) retrieval. In this experiment, we evaluated the retrieval performance of StickerLLM on the Sticker820K test set. 

\begin{table}[htbp]
\centering
\setlength\tabcolsep{3pt}
\begin{tabular}{ccccccc} 
\toprule
Method                      & Params                & Modality & R@1  & R@5  & R@10 & MR    \\ 
\hline
\multirow{2}{*}{StickerLLM} & \multirow{2}{*}{3M} & I2I      & 96.9 & 99.9 & 99.9 & 99.0  \\
                            &                       & T2I      & 61.5 & 82.4 & 87.5 & 77.1  \\
\bottomrule
\end{tabular}
\caption{Experimental results of StickerLLM in image-to-image (I2I) retrieval and text-to-image (T2I) retrieval.}
\label{tab:sticker_llm_eval}
\vspace{-0.4cm}
\end{table}

As shown in Tab. \ref{tab:sticker_llm_eval}, the performance of StickerLLM to retrieve stickers based on images is much higher than its ability to retrieve stickers based on text (96.9\% R@1 image-to-image retrieval vs. 61.5\% text-to-image  
 retrieval). This is mainly because image features have richer and more detailed semantic information compared to text features. Combining Tab. \ref{tab:sticker_clip_eval}, when using ViT-B/16 as the image encoder, StickerLLM's performance on text-to-image retrieval exceeds the freeze-vision version based on ViT-B CLIP (77.1\% MR vs. 67.9\% MR), and is only surpassed by StickerCLIP ViT-B (77.1\% MR vs. 82.7\% MR). At the same time, StickerLLM's amount of trainable parameters is much lower than these two models, which demonstrates the potential for expanding the capabilities of LLM through prompt tuning.

\subsection{Qualitative Results}
\begin{figure}
\centering
\includegraphics[width=0.47\textwidth]{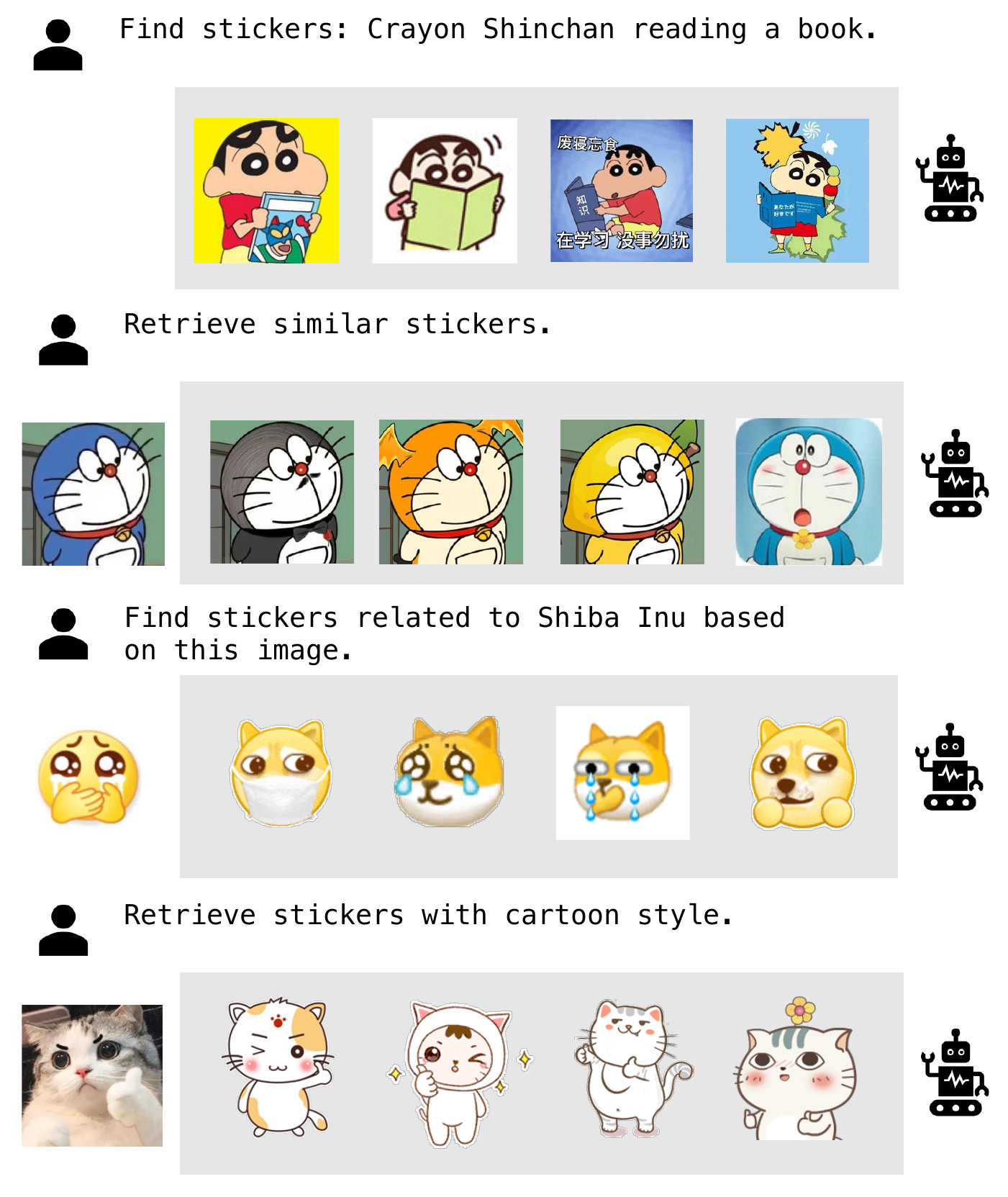}
\caption{Visualization results of StickerLLM on Sticker820K.}
\label{fig:stickerllm_example}
\vspace{-0.6cm}
\end{figure}

\textbf{Visualization on sticker retrieval.} StickerLLM accepts text queries and images as input, utilizes $\langle ret \rangle$ token to extract a unified retrieval embedding, which is aligned with the output features of visual encoder in StickerCLIP during the training process. In Fig. \ref{fig:stickerllm_example}, we present the visualization results of StickerLLM on the Sticker820K dataset. We show StickerLLM can not only search stickers based on text queries alone, but also combine text and images for more refined retrieval. By randomly adding images as noise with a certain probability in the image-text instruction training data, StickerLLM can focus on text information rather than relying entirely on image information during joint image-text retrieval. The visual results of joint text and image retrieval by StickerLLM are shown in the third and fourth rows of Fig. \ref{fig:stickerllm_example}, where it can be observed that the search results of StickerLLM also meet the requirements of text queries when they are similar to the input images.

\begin{figure}
\centering
\includegraphics[width=0.47\textwidth]{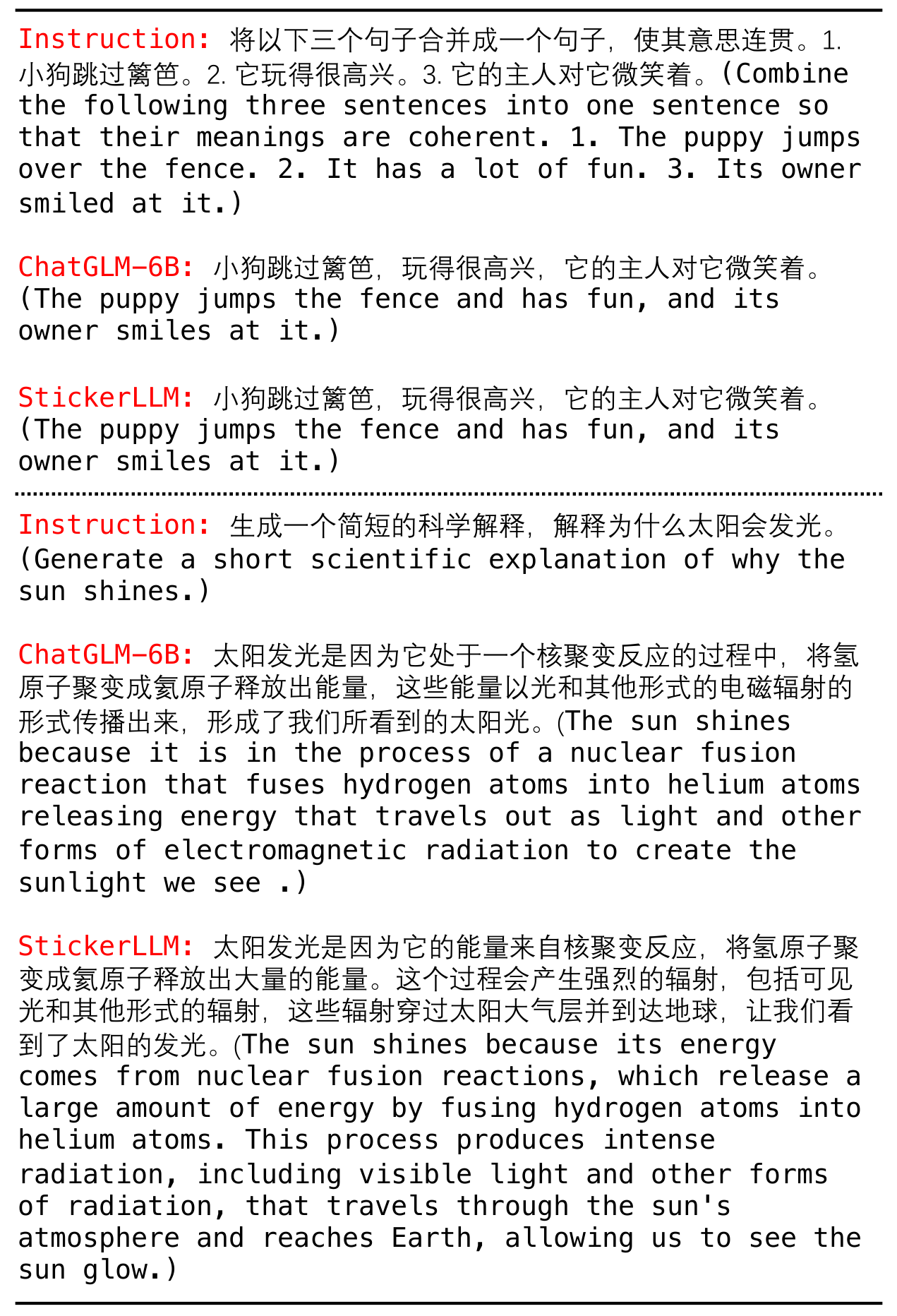}
\caption{Qualitative comparison between StickerLLM and ChatGLM-6B. We only fine-tuned the embedding of the added special tokens in LLM, hence the answers are similar for the same question.}
\label{fig:stickerllm_vs_chatglm}
\vspace{-0.4cm}
\end{figure}

\textbf{Visualization on common tasks.} We utilize ChatGLM-6B as the pretrained model of StickerLLM, freeze the model parameters and update only the embeddings corresponding to the positions of the added special tokens and linear layers $\mathbf{W}_t$, $\mathbf{W}_c$. Thus it will not impact the performance of LLM on common tasks. In Fig. \ref{fig:stickerllm_vs_chatglm}, we showcase the responses of StickerLLM and ChatGLM-6b on common instructions. It can be seen that the two models provide almost identical answers for the same instruction.

%% file: 6-conclusion.tex
\section{Conclusion}
In this work, we have introduced a vision-language dataset concerning stickers, namely Sticker820K. To the best of our knowledge, this is the first Chinese large-scale image-text pair dataset for stickers. Our dataset has been enriched through manual text annotations of stickers that encompass content descriptions, emotional labels, style labels and OCR transcripts. Based on the Sticker820K, we have released two benchmark models, namely StickerCLIP and StickerLLM. In the experiments of StickerCLIP, we have demonstrated the necessity of finetuning CLIP through comparison with zero-shot models, in the task of aligning image-text pairs of stickers. As for StickerLLM, we focus on the latest widely studied LLM, and have endeavored to stretch its multimodal capabilities on the task of sticker retrieval. We have incorporated novel tokens as tools, and exclusively fine-tuned the embedding and mapping of these new tokens. Hence, there is a hardware friendly way for fine-tuning LLM, and its performance under other instructions will remain unscathed. 